\documentclass[acmsmall,screen]{acmart}

\usepackage{multirow} 

\AtBeginDocument{%
  \providecommand\BibTeX{{%
    \normalfont B\kern-0.5em{\scshape i\kern-0.25em b}\kern-0.8em\TeX}}}

\begin{document}

\title{Prior Knowledge Distillation Network for Face Super-Resolution}

\author{Qiu Yang}
\affiliation{%
  \institution{School of Computer and Information, Hefei University of Technology}
  \city{Hefei}
  \country{China}}
\email{2022171284@mail.hfut.edu.cn}

\author{Xiao Sun}
\authornote{Corresponding authors}
\affiliation{%
  \institution{School of Computer and Information, Hefei University of Technology}
  \city{Hefei}
  \country{China}}
\email{sunx@hfut.edu.cn}

\author{Xin-yu Li}
\affiliation{%
  \institution{School of Computer and Information, Hefei University of Technology}
  \city{Hefei}
  \country{China}}
\email{2022171284@mail.hfut.edu.cn}

\author{Feng-Qi Cui}
\affiliation{%
  \institution{Institute of Advanced Technology, University of Science and Technology of China}
  \city{Hefei}
  \country{China}}
\email{fengqi_cui@mail.ustc.edu.cn}

\author{Yu-Tong Guo}
\affiliation{%
  \institution{School of Computer and Information, Hefei University of Technology}
  \city{Hefei}
  \country{China}}
\email{hfutgyt@163.com}

\author{Shuang-Zhen Hu}
\affiliation{%
  \institution{School of Computer and Information, Hefei University of Technology}
  \city{Hefei}
  \country{China}}
\email{hfut_shuangzhen@qq.com}

\author{Ping Luo}
\affiliation{%
  \institution{School of Computer and Information, Hefei University of Technology}
  \city{Hefei}
  \country{China}}
\email{2022212392@mail.hfut.edu.cn}

\author{Si-Ying Li}
\affiliation{%
  \institution{School of Computer and Information, Hefei University of Technology}
  \city{Hefei}
  \country{China}}
\email{hfut_siying@163.com}

\renewcommand{\shortauthors}{Qiu Yang and Xiao Sun, et al.}

\begin{abstract}
The purpose of face super-resolution (FSR) is to reconstruct high-resolution (HR) face images from low-resolution (LR) inputs. With the continuous advancement of deep learning technologies, contemporary prior-guided FSR methods initially estimate facial priors and then use this information to assist in the super-resolution reconstruction process. However, ensuring the accuracy of prior estimation remains challenging, and straightforward cascading and convolutional operations often fail to fully leverage prior knowledge. Inaccurate or insufficiently utilized prior information inevitably degrades FSR performance. To address this issue, we propose a prior knowledge distillation network (PKDN) for FSR, which involves transferring prior information from the teacher network to the student network. This approach enables the network to learn priors during the training stage while relying solely on low-resolution facial images during the testing stage, thus mitigating the adverse effects of prior estimation inaccuracies. Additionally, we incorporate robust attention mechanisms to design a parsing map fusion block that effectively utilizes prior information. To prevent feature loss, we retain multi-scale features during the feature extraction stage and employ them in the subsequent super-resolution reconstruction process. Experimental results on benchmark datasets demonstrate that our PKDN approach surpasses existing FSR methods in generating high-quality face images.
\end{abstract}



\maketitle

\section{Introduction}
Face super-resolution, commonly referred to as face hallucination, seeks to produce HR face images from their LR counterparts. In practical scenarios, image resolution is often compromised due to various factors, including the quality of mobile video devices, natural weather conditions, and human activities, leading to reduced resolution, blurriness, and distortion in the resulting images. Consequently, the importance of face super-resolution reconstruction is evident, as it plays a critical role in facial recognition, expression recognition, and attribute analysis, among other face-related image processing applications.

Recent advancements in deep learning and substantial gains in computational power have facilitated the emergence of numerous effective deep learning-based methods for FSR \cite{wang2023spatial, he2022gcfsr, wang2023super, kim2016accurate, wang2024facialpulse}, demonstrating notable improvements. Given that faces are highly structured objects with specific facial features and identity information, researchers have focused on leveraging facial priors, such as landmarks, heatmaps, and parsing map, to enhance the reconstruction of high-resolution face images. 
Methods guided by these priors generally begin by estimating the facial priors first and then use these priors to assist in producing high-quality face images, thereby improving FSR performance.

Despite advancements in prior-guided FSR methods, challenges remain. The accuracy of facial priors, which can be compromised by errors in priors derived from low-resolution images, may be further exacerbated in subsequent iterations, significantly impacting FSR performance. Additionally, straightforward cascading operations often fail to fully leverage the extracted priors, thus constraining potential improvements in super-resolution performance.

To address these challenges, inspired by KDFSRNet \cite{wang2022propagating}, we introduce a prior knowledge distillation approach for the FSR task. In our approach, the teacher network utilizes facial parsing map obtained from ground truth high-resolution images to restore high-quality faces, while the student network carries out super-resolution reconstruction using only LR face inputs under the guidance of the teacher network. This method allows the teacher network to transfer prior knowledge to the student network, mitigating the adverse effects of inaccurate prior estimation. Moreover, to fully utilize facial prior information, we design a parsing map fusion block (PFB) that explores information dependencies across spatial and channel dimensions and integrates facial parsing map, thus enhancing the utilization of prior information. To prevent the loss of complementary features, we have incorporated a feature fusion block (FFB), which retains more details and contextual information during the super-resolution reconstruction process. The key contributions of this study are outlined as follows:
\begin{itemize}
    \item Development of a prior knowledge distillation network (PKDN) for face super-resolution that operates without prior estimation. 
    \item Introduction of the parsing map fusion block to further integrate prior knowledge with inter-channel and spatial relationships, promoting their synergy and enhancing FSR performance. 
    \item Experiments conducted on the widely used benchmark datasets CelebA \cite{liu2015deep} and Helen \cite{le2012interactive} reveal that our method outperforms existing FSR techniques in generating high-fidelity super-resolved face images.
\end{itemize}

\begin{figure*}[t]
	\centering
	\includegraphics[width=1.0\textwidth]{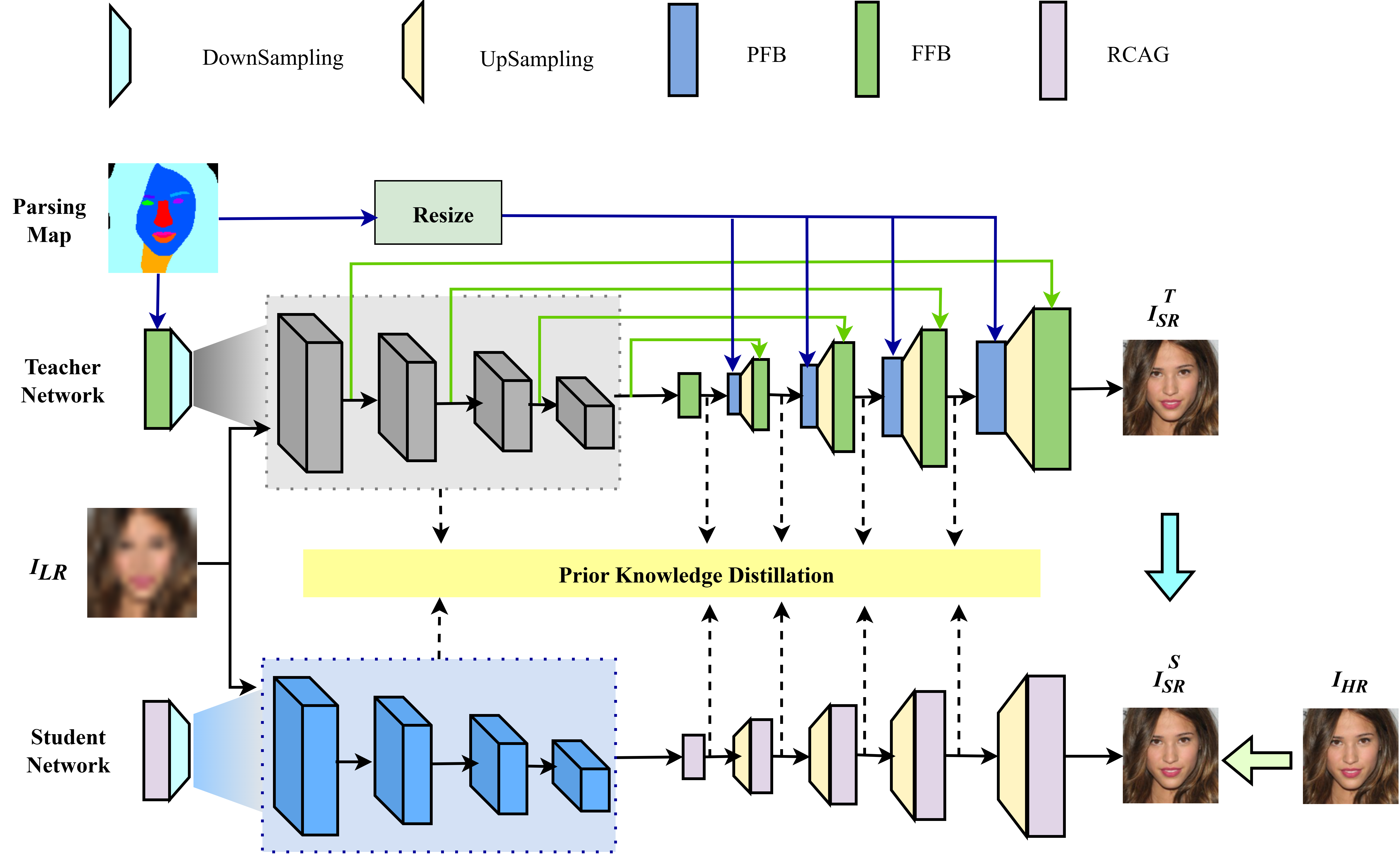} %
	\caption{The network architecture of our proposed PKDN.}
    \Description{fig:PKDN}
	\label{fig:PKDN}
\end{figure*}

\section{Related Work}

\subsection{General FSR Methods in Deep Learning}
As deep learning technologies advance rapidly, numerous neural network-based FSR methods emerge \cite{wang2023spatial, he2022gcfsr,zhang2018residual,li2022counterfactual, huang2023phyfinatt, yang2020hifacegan, ai2024ais,gu2023wife}, achieving remarkable results. In the early stage, researchers employed convolutional neural networks (CNNS) for FSR reconstruction without incorporating additional facial prior information. These methods are referred to as generic FSR approaches. Huang et al. \cite{huang2017densely} develope DenseNet, which employs dense connections between feature blocks within super-resolution neural networks, facilitating more efficient learning and improved model performance. Wang et al. \cite{wang2023spatial} develope an FSR method called SFMNet, which utilizes Fourier Transform to investigate the relationships between the spatial and frequency domains. Recently, with the rise of attention mechanisms, many researchers have focused on developing effective attention mechanisms. For instance, SPARNet \cite{chen2020learning} introduce a spatial attention residual network, incorporating facial attention units and combining spatial attention networks to capture crucial facial structures. He et al.\cite{he2022gcfsr} present a generative controllable FSR framework called GCFSR, which can produce realistic facial details. This framework effectively integrates multi-layer encoded features with generated features dynamically, achieving impressive results.

\subsection{Prior-guided FSR Methods}
As FSR is a domain-specific super-resolution method with facial images being highly structured objects possessing relatively fixed structural information and prior knowledge, leveraging facial priors can significantly enhance reconstruction performance. Many prior-based FSR methods \cite{chen2018fsrnet, yu2018face, ma2020deep, wang2023super, huang2024keystrokesniffer} have achieved impressive milestones. FSRNet \cite{chen2018fsrnet} proposes an end-to-end FSR method that first predicts facial landmarks and parsing map, which are then utilized to recover high-quality facial images. DIC \cite{ma2020deep} develope an attention fusion module to exploit facial priors, iteratively performing FSR and prior estimation to boost overall model performance. KDFSRNet \cite{wang2022propagating} design a prior knowledge distillation framework to utilize facial parsing map for clearer facial image reconstruction. Although these methods produce impressive results, the accuracy of prior estimation remains challenging, and the utilization of prior information is often inadequate, inevitably limiting the overall reconstruction quality of the models.

\section{Method}
As illustrated in Figure \ref{fig:PKDN}, our proposed PKDN features a robust teacher autoencoder network and a streamlined student network with a comparable structure. The teacher network, which incorporates a parsing map fusion block to effectively utilize prior information, directly extracts facial parsing maps from ground truth HR images. These prior information are then transferred to the student network via a knowledge distillation strategy. To further enhance performance, we introduce a feature fusion block to integrate prior low-resolution information, mitigating the risk of losing complementary features. This approach helps the student network learn prior knowledge effectively while minimizing the impact of inaccurate prior estimates on super-resolution reconstruction.

\begin{figure}[t]
\centering
\includegraphics[scale=0.6]{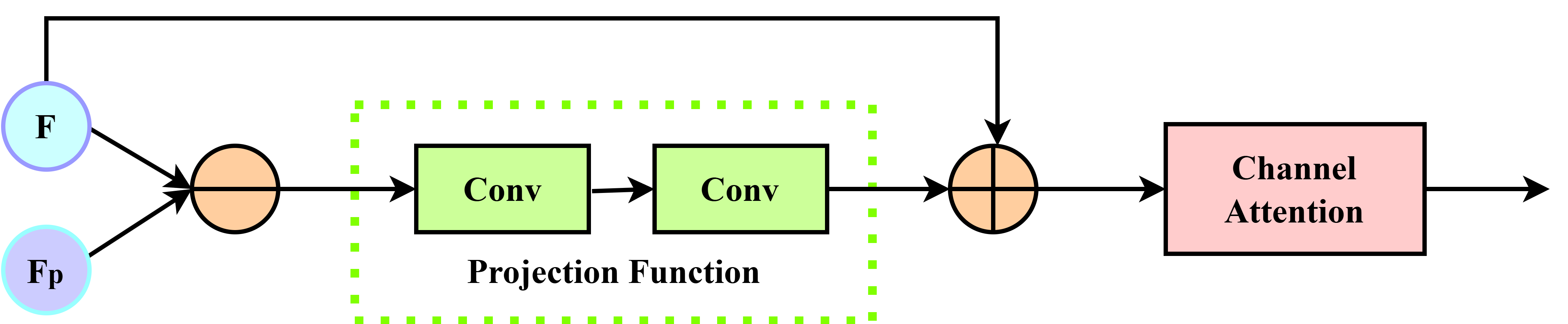}
\caption{The architecture of our feature fusion block (FFB). We use previously retained feature $F_{p}$ with the same resolution as the current feature F for feature fusion, which compensates for the loss of complementary features in the super-resolution reconstruction process. We first calculate the difference between F and $F_{p}$, and then process this difference through an error feedback mechanism called the Projection Function \cite{wang2023super}. The processed difference is then added back to F and refined using a channel attention mechanism to preserve more details and enhance feature representation.}
\label{fig:FFB}
\Description{fig:FFB}
\end{figure}

\subsection{Teacher Network}\label{AA}
To fully leverage facial prior knowledge, the teacher network takes as input both LR face images and facial parsing maps extracted from the corresponding ground truth HR face images. The encoder consists of downsampling blocks and parsing map fusion blocks. The downsampling block uses inverse pixel-shuffle to halve the feature resolution. The combination of upsampling blocks, parsing map fusion blocks and feature fusion blocks constitutes the decoder of teacher network. HR features typically encompass rich details and finer texture information, while LR features provide abundant contextual information. To prevent the loss of these complementary features during the super-resolution reconstruction process, we retain multi-scale features from the feature extraction phase and utilize them in subsequent upsampling stages. Specifically, we modified the multi-scale refine block in FishFSRNet \cite{wang2023super} to design a feature fusion block (FFB), as illustrated in Figure \ref{fig:FFB}. To accelerate model training, we use prior features at the same resolution as the current features for feature fusion. Our upsampling block utilizes pixel-shuffle to enhance feature resolution. The loss function employs $\mathcal{L}_1$ loss, defined as follows:
\begin{equation}
    \mathcal{L_{T} }  = \parallel I_{SR}^{T} -I_{HR}  \parallel _{1} 
\end{equation}
where $\mathcal{L_{T} }$ represents the loss function of the teacher network, $I_{SR}^{T}$ denotes the SR face generated by the teacher network, $I_{HR}$ signifies the ground truth.

\subsection{Student Network}
In the training phase, the teacher network leverages HR facial parsing map, which allows it to fully utilize prior information\cite{liu2023finch}. However, during the inference stage, such HR prior information is unavailable. To address this, the knowledge distillation strategy enables the student network to effectively utilize facial prior information even when only LR face images are provided as input. This approach helps mitigate the adverse effects that arise from inaccurate prior estimations. The student network employs an auto-encoder architecture, comprising upsampling blocks, downsampling blocks, and residual channel attention group (RCAG) \cite{wang2022propagating} with multiple cascaded RCABs \cite{zhang2018image}. The RCAG is tailored to effectively capture the characteristics of various layers. The student network receives only LR face images as input. The overall loss function of the student network $\mathcal{L_{S} }$ is formuated as Equation \ref{equation:overall}.
\begin{equation}
    \mathcal{L_{S} } = \mathcal{L_{SR} } + \lambda _{TS}\mathcal{L_{TS} } + \lambda _{FS} \mathcal{L_{FS} }
    \label{equation:overall}
\end{equation}
where $\lambda _{TS}$ and $\lambda _{FS}$ are the penalty coefficients of the $\mathcal{L_{TS} }$ and $\mathcal{L_{FS} }$ to balance the loss.
\begin{equation}
    \mathcal{L_{SR} } = \parallel I_{SR}^{S} - I_{HR}  \parallel _{1}
\end{equation}
where $I_{SR}^{S}$ represents the SR face produced by the student network, $\mathcal{L_{SR} }$ is the pixel-wise $\mathcal{L}_1$ loss function computed between $I_{HR}$ and $I_{SR}^{S}$.
\begin{equation}
    \mathcal{L_{TS} } = \parallel I_{SR}^{T} - I_{SR}^{S} \parallel _{1}
\end{equation}
where $\mathcal{L_{TS}}$ enhances the student network's capacity to assimilate and integrate prior knowledge from the teacher network\cite{liu2024federated}. Furthermore, $\mathcal{L_{TS}}$ provides the student network with more accessible learning signals compared to the ground truth and includes critical information gleaned by the robust teacher network.

\begin{equation}
    \mathcal{L_{FS} } =  \frac{1}{n} \sum_{i}^{n}  \parallel F_{i}^{T} - F_{i}^{S} \parallel_{1}
\end{equation}
where $F_{i}^{T}$ and $F_{i}^{S}$ denote the $i$-th intermediate features from the teacher and student networks, respectively. $\mathcal{L}_{FS}$ specifically imposes constraints on these features, which are derived from each downsampling and upsampling block.

\subsection{Parsing map Fusion Block}
\begin{figure}[t]
\centering
\includegraphics[scale=0.6]{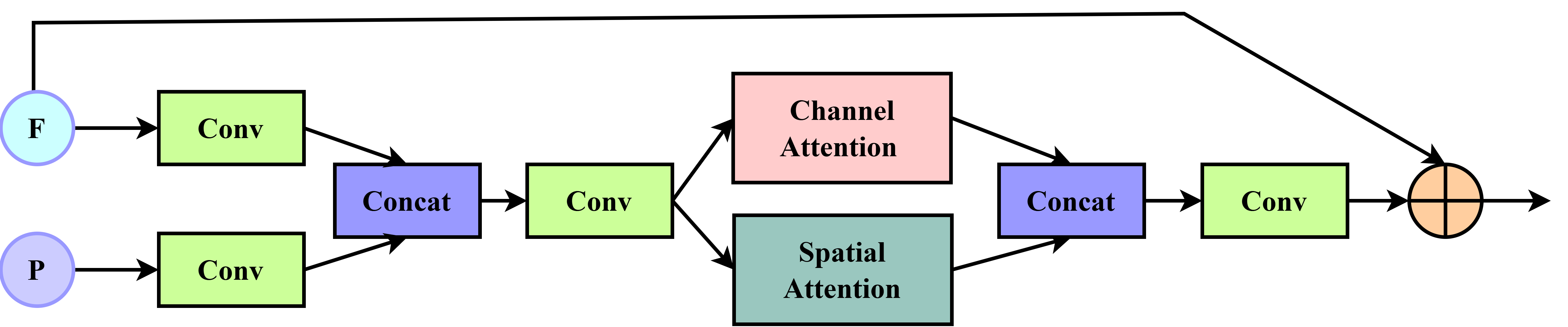}
\caption{The architecture of our paring map fusion block (PFB). We incorporated spatial and channel attention mechanisms into the parsing map fusion block to fully leverage facial parsing map.}
\Description{fig:PFB}
\label{fig:PFB}
\end{figure}

By integrating facial feature information with an robust attention mechanism, we develope a parsing map fusion block, as illustrated in Figure \ref{fig:PFB}. The proposed PFB effectively captures information dependencies across both spatial and channel attention dimensions, while also fully utilizing the parsing map. Specifically, we insert parsing map P into a value of the same size as current feature F through recent interpolation. Two convolutional layers are then applied to project them into similar domains to extract diverse information. A fusion convolution layer is added after concatenating the projection features. These features are subsequently refined using parallel channel and spatial attention mechanism to capture dependencies across channels and spatial dimensions. To integrate these information effectively, we directly connect their outputs and apply a fusion convolution layer for further combination. Finally, the resulting features are combined with F to enhance super-resolution performance.

\section{Experiment}

\subsection{Datasets and Implementation}
We train our PKDN using the CelebA \cite{liu2015deep} and Helen \cite{le2012interactive} datasets. For the CelebA dataset, we adopt the experimental setup from \cite{ma2020deep}, selecting 168,854 face images for training, with 100 images allocated for performance verification during training and 1,000 images reserved for the test phase. Similarly, for the Helen dataset, we also followe the experimental setup in \cite{ma2020deep}: using 1955 face images for training, 50 face images for verification and 50 as the test set. We resize face images to 128 × 128 pixels to establish the ground truth. Subsequently, we downsampling these ground truth images to 16 × 16 pixels to generate LR face images for ×8 FSR. For the parsing map, we followed the experimental setup of KDFSRNet \cite{wang2022propagating}, using pre-trained BiSeNet \cite{yu2018bisenet} to process HR faces and obtain the ground truth parsing map. We implement our PKDN using PyTorch on a GeForce RTX 3090 GPU.

\subsection{Comparisons with State-of-the-Art Methods}
We compare our method with several advanced FSR methods on upgrade factor ×8, including three general FSR methods URDGN \cite{yu2016ultra}, SPARNet \cite{chen2020learning}, SISN \cite{lu2021face} and three FSR methods guided by facial prior knowledge including FSRNet \cite{chen2018fsrnet}, DIC \cite{ma2020deep}, KDFSRNet \cite{wang2022propagating}.

\setlength{\tabcolsep}{6pt} 
\renewcommand{\arraystretch}{1.0} 
\begin{table}[t]
    \centering
    \caption{Quantitative comparisons with leading approaches on the CelebA \cite{liu2015deep} and Helen \cite{le2012interactive} datasets for a scale factor of 8.}
    \begin{tabular}{c|cc|cc}
    \toprule
        \multirow{2}{*}{Method} & \multicolumn{2}{c}{CelebA} & \multicolumn{2}{c}{Helen} \\ \cmidrule{2-5}
        & PSNR & SSIM & PSNR & SSIM \\ \midrule
        URDGN \cite{yu2016ultra} & 25.62 & 0.7261 & 25.23 & 0.7205  \\ 
        FSRNet \cite{chen2018fsrnet} & 26.66 & 0.7714 & 26.28 & 0.7726  \\ 
        DIC \cite{ma2020deep} & 27.27 & 0.8022 & 26.69 & 0.7953  \\ 
        SISN \cite{lu2021face} & 27.31 & 0.7978 & 26.66 & 0.7920  \\ 
        SPARNet \cite{chen2020learning} & 27.42 & 0.8036 & 26.78 & 0.7971  \\ 
        KDFSRNet \cite{wang2022propagating} & 27.52 & 0.8057 & 26.79 & 0.7953  \\ 
        PKDN & \textbf{27.56} & \textbf{0.8062} & \textbf{26.84} & \textbf{0.7982} \\ \bottomrule
    \end{tabular}
    \label{table:all}
\end{table}

\begin{figure}[t]
	\centering
	\includegraphics[width=1.0\textwidth]{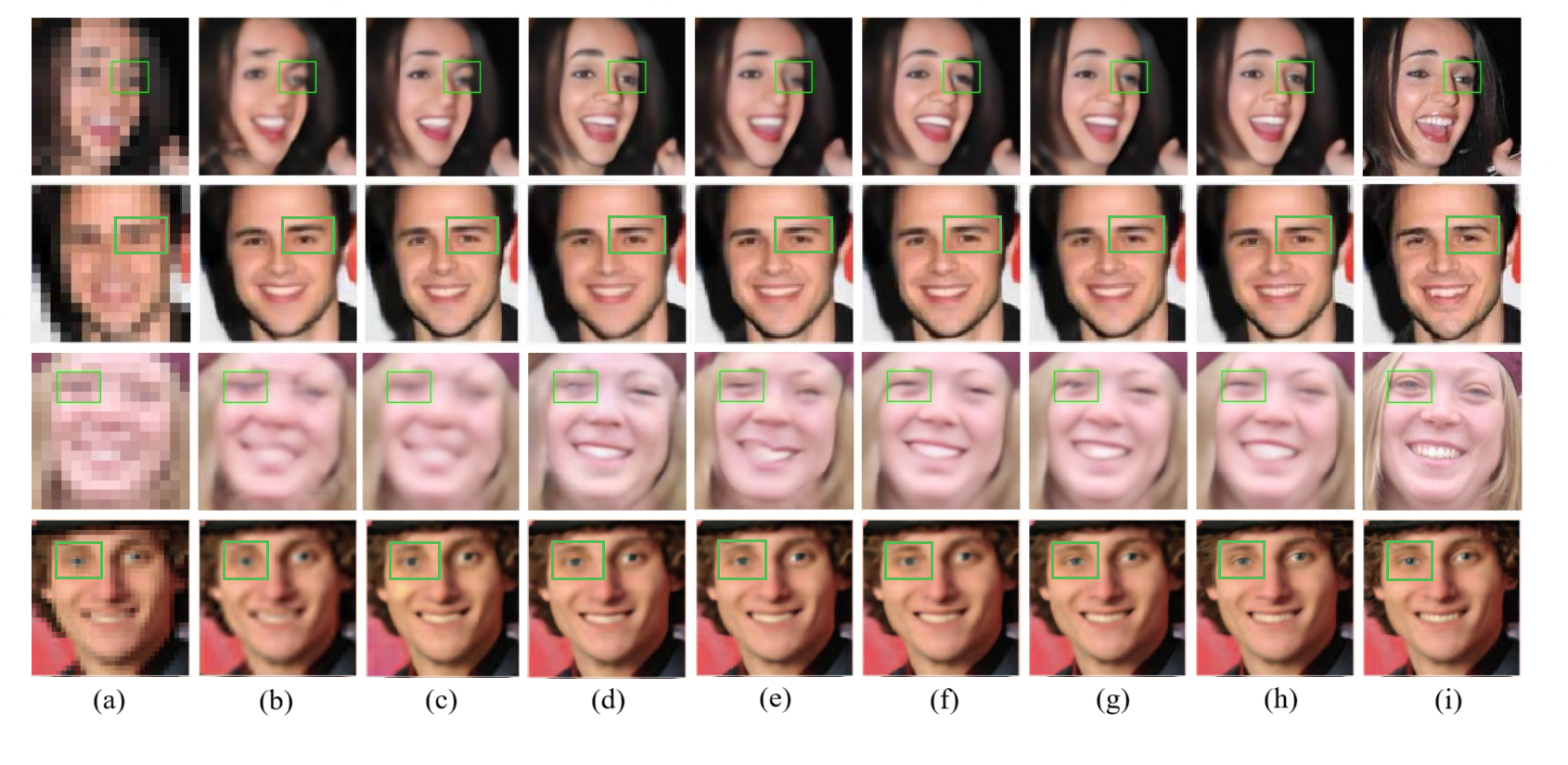} %
	\caption{Visual quality comparison of leading methods on CelebA \cite{liu2015deep} dataset (top two rows) and Helen \cite{le2012interactive} dataset (bottom two rows) for a scale factor of 8. (a): LR; (b): URDGN \cite{yu2016ultra}; (c): SPARNet \cite{chen2020learning}; (d): SISN \cite{lu2021face}; (e): FSRNet \cite{chen2018fsrnet}; (f): DIC \cite{ma2020deep}; (g): KDFSRNet \cite{wang2022propagating}; (h): Ours; (i): HR.}
    \Description{fig:visual}
	\label{fig:visual}
\end{figure}

Quantitative results are shown in Table \ref{table:all}. Our PKDN outperforms existing methods in both PSNR and SSIM metrics. Compared to general FSR methods \cite{yu2016ultra, chen2020learning, lu2021face}, our approach leverages facial prior information more effectively through the PFB, achieving better results. In comparison to prior-guided FSR methods \cite{chen2018fsrnet, ma2020deep, wang2022propagating}, our PKDN significantly reduces the adverse effects of inaccurate prior estimation through knowledge distillation, resulting in superior performance. Figure \ref{fig:visual} presents the qualitative results, showcasing the visual outputs of various methods with an upscaling factor of ×8 on the CelebA dataset. Leveraging facial prior knowledge via the PFB, our method generates contours and shapes with greater clarity. Compared to other methods, our PKDN achieves superior reconstruction of facial textures, providing visually enhanced results.

\subsection{Ablation study}
In this part, we further validate the effectiveness of the key components PFB and FFB in PKDN through experiments on the CelebA dataset. For a clear comparison, we first replace the PFB module in the teacher network with the RCAG used in the student network and remove the FFB module. Next, we remove the FFB module from the PKDN model while retaining the carefully designed PFB module. Finally, we include both the PFB and MSRB modules in the model, which we refer to as our complete PKDN model. The results presented in Table \ref{table:albation}, indicate that the designed and introduced modules significantly enhance FSR performance.
\setlength{\tabcolsep}{6pt} 
\renewcommand{\arraystretch}{1.0} 
\begin{table}[t]
    \centering
    \caption{Research on the effectiveness of different components of the proposed network on CelebA \cite{liu2015deep} with a scale of ×8. PFB and FFB represent the parsing map fusion block and feature fusion block, respectively.}
    \begin{tabular}{c|cc|cc}
    \toprule
        Model & PFB & FFB & PSNR & SSIM  \\ \hline
        without PFB and FFB &  W/O &  W/O & 27.31 & 0.7961  \\ 
        without PFB &  W/ &  W/O & 27.48 & 0.8036  \\ 
        PKDN & W/ & W/ & \textbf{27.56} & \textbf{0.8062} \\ \bottomrule
    \end{tabular}
    \label{table:albation}
\end{table}
\section{Conclusion}
In this paper, we introduce a facial Prior Knowledge Distillation Network for FSR. Drawing inspiration from KDFSRNet \cite{wang2022propagating}, our PKDN architecture closely follows that of KDFSRNet but incorporates several novel elements. Specifically, we design a parsing map fusion block to more efficiently leverage facial prior information and explore dependencies along both spatial and channel attention dimensions. Additionally, we introduce a feature fusion block to fully utilize features across different resolutions. Extensive experiments show that our method delivers improved performance.

\bibliographystyle{ACM-Reference-Format}
\bibliography{acmart}

\end{document}